\documentclass[conference]{IEEEtran}
\IEEEoverridecommandlockouts
\newif\ifanon
\anonfalse

\usepackage{cite}
\usepackage{amsmath,amssymb,amsfonts}
\usepackage{graphicx}
\usepackage{booktabs}
\usepackage{multirow}
\usepackage{array}
\usepackage{url}
\usepackage[hidelinks]{hyperref}
\usepackage{textcomp}
\usepackage{xcolor}
\usepackage[shortlabels]{enumitem}
\def\BibTeX{{\rm B\kern-.05em{\sc i\kern-.025em b}\kern-.08em T\kern-.1667em\lower.7ex\hbox{E}\kern-.125emX}}
\newcommand{\figref}[1]{Fig.~\ref{#1}}
\newcommand{\tabref}[1]{Table~\ref{#1}}
\newcommand{\secref}[1]{Section~\ref{#1}}
\begin{document}

\title{Alert-List Concordance for Evaluating Imputation Strategies in ICU Sepsis Prediction}

\ifanon
\author{\IEEEauthorblockN{Anonymous Authors}
\IEEEauthorblockA{Anonymous submission}}
\else
\author{
\IEEEauthorblockN{Priyanka Siddappa, Muntaqim Ahmed Raju, Shifat Haider Al Amin, and Ruizhe Ma}
\IEEEauthorblockA{Miner School of Computer \& Information Sciences, University of Massachusetts Lowell, Lowell, MA 01854, USA\\
\{Priyanka\_Siddappa, MuntaqimAhmed\_Raju, MdShifat\_AlAmin\}@student.uml.edu, Ruizhe\_Ma@uml.edu\\
}
}
\fi

\maketitle
\bstctlcite{IEEEexample:BSTcontrol}

\begin{abstract}
Preprocessing is an under-measured reliability risk in clinical prediction: a model can keep a similar AUROC and still change who receives an alert. Imputation quality is commonly evaluated using reconstruction error or aggregate discrimination, neither of which directly measures agreement on which patients are flagged. At a matched alert rate, pipelines with similar aggregate performance can produce different patient lists. This question is especially relevant under the frequent, structured missingness of real ICU records. We therefore introduce alert-list concordance, which asks whether two pipelines identify the same high-risk patients, measured by the Jaccard similarity of their top-$k$\% alert sets with patient-level bootstrap confidence intervals. On PhysioNet/CinC 2019 (40,336 patients, 71.9\% missingness), we compare nine imputation strategies and a native-missingness baseline using a fixed XGBoost configuration, fitted and Platt-calibrated separately for each strategy. After Bonferroni correction, 22 of the 28 reconstruction-method pairs showed no statistically significant difference in patient-level AUROC, yet their alert lists overlap at Jaccard 0.518 to 0.750. Similar aggregate discrimination can therefore accompany different patient selections. ForwardFill is a separate outlier, pairing higher discrimination with the lowest overlap in the matrix, 0.40 to 0.45. The broad pattern holds within each hospital system and under six patient-level aggregation rules. At the row level, eight of ten methods fall within an AUROC range of 0.011. In this row-level pipeline, deep imputers did not outperform mean imputation on discrimination, and the sequential implementations cost 5 to 12 times mean imputation in total pipeline wall-clock time on our hardware.
\end{abstract}


\section{Introduction}
\label{sec:intro}

Missing data are the normal condition of ICU records rather than an exception. Tests are ordered when clinicians judge them necessary, monitors are disconnected, and some analytes are measured only in specific subpopulations. In the PhysioNet/CinC 2019 cohort, labeled by Sepsis-3 criteria~\cite{singer_sepsis3_2016}, median absence exceeds 90\% for laboratory variables (\secref{sec:data}). Observation patterns can be informative because the probability that a value is recorded is associated with disease severity~\cite{tan2023informative}. This association alone does not distinguish missing-at-random from missing-not-at-random mechanisms in Rubin's taxonomy~\cite{rubin1976inference}. Many classifiers require missing values to be imputed before fitting. Several approaches are available: mean or last-observation-carried-forward substitution, chained equations~\cite{buuren_mice_2018}, denoising and masked autoencoders~\cite{vincent_denoising_autoencoders_2010,he_mae_2022,gondara_mida_2018}, and sequential deep imputers~\cite{du_saits_2023,cao2018brits,che_grud_2018}. Tree learners can absorb missingness natively~\cite{chen_xgboost_2016}. The choice can matter because reconstruction toward a population distribution may replace clinically informative absence with values closer to the population average, whereas temporal propagation carries the most recent, possibly deteriorating, observation forward.

In practice the choice is made on aggregate metrics, reconstruction error~\cite{luo_dacmi_2022,poette2026benchmarking} or discrimination~\cite{perezlebel_missing_values_benchmark_2022}, and neither is aligned with deployment. A sepsis early-warning system does not act on a ranking. It issues alerts to a limited share of the census, 18\% of hospitalized patients for one widely deployed model~\cite{wong_epic_sepsis_2021}, so its operative output is a \emph{set} of flagged patients. The practical stake is small and specific: in a 100-bed unit alerting on ten patients, how many of those ten change when only the imputation step changes? An average over all patients cannot reveal whether two pipelines flag the same set, a divergence that would carry into outcome studies and equity audits alike~\cite{getzen2023equitable}. These aggregate metrics do not directly establish whether pipelines flag the same patients, so the question goes untested in the regime clinical data actually occupy.

We therefore ask a different question: does imputation choice change \emph{which patients} are flagged? Our contributions are summarized as follows:
\begin{enumerate}[leftmargin=*,itemsep=1pt,topsep=2pt]
\item We introduce alert-list concordance, which compares imputation pipelines by the Jaccard overlap of the top-$k$\% high-risk sets they produce at the institution's alert rate, with patient-level bootstrap confidence intervals;
\item We show that the resulting concordance structure is tiered and that ForwardFill is a structural outlier, with an alert-list Jaccard of only 0.40 to 0.45 against every other method, and that the tiering survives site-wise and aggregation-rule checks;
\item We show that AUROC is nearly blind to this divergence, with eight of ten methods inside 0.011, so aggregate equivalence does not imply operational equivalence;
\item We characterize the patients on whom the strategies disagree, and report operating-point utility, calibration and compute cost alongside concordance.
\end{enumerate}

\section{Related Work}
\label{sec:related}

A retrospective external validation of a widely used proprietary
sepsis model found that the evaluated threshold would flag 18\%
of hospitalizations while missing 67\% of sepsis cases
before onset~\cite{wong_epic_sepsis_2021}. On the separate
PhysioNet/CinC 2019 cohort, Gupta et al.\ investigated reducing
false alarms in sepsis prediction~\cite{gupta_sepsisai_2024}.
These studies highlight the importance of evaluating alert burden
and detection performance. We examine a complementary question:
how imputation choice changes which patients are flagged.

This matters because in the ICU an absent measurement is itself a signal, so whatever an imputer writes into a gap either preserves that signal or erases it. Sensor disconnection removes contiguous blocks of a variable, a structure that benchmarks based on random deletion systematically under-penalize~\cite{poette2026benchmarking}. Selective test ordering can produce observation patterns that carry prognostic signal: missingness indicators alone, with no measured values, reach an AUROC of 0.68 for 30-day mortality in MIMIC-III~\cite{sharafoddini_missing_icu_2019}, and ordering patterns track disease severity~\cite{tan2023informative}. The established remedy is to expose that pattern to the classifier directly, as indicators or as time since last observation~\cite{lipton2016missing,che_grud_2018,vanness2023mim,sisk2023indicators}. We omit them by design, so that the values written into the gaps remain the only quantity varying across the comparison (\secref{sec:limits}).

Which imputer to use is then decided empirically, and the evidence is mixed. Simulation on an ICU registry favors multiple imputation over ad hoc rules~\cite{steif_imputation_sim_icu_2022}, while comparisons of deep imputers find that larger models do not necessarily help~\cite{qian2024howdeep} and that simple interpolation often comes close~\cite{poette2026benchmarking}. Multi-center benchmarks now make such comparisons reproducible~\cite{vandewater2023yaib}, and one finds that preprocessing moves prediction performance often more than model class does~\cite{vandewater2023yaib}. A newer generation tries to keep informative absence rather than overwrite it, by modelling the mechanism~\cite{qian2023csai} or by refusing to impute at all~\cite{neog2026misstsm}. They are still scored the same way.

That aggregate discrimination is an insufficient criterion is itself long established. Cook argued nearly two decades ago that it is the wrong lens for prognostic models~\cite{cook_beyond_roc_2008}. Calibration and clinical usefulness have been advanced as complements~\cite{walsh_beyond_discrimination_2017}, and threshold analyses show that clinically optimal operating points diverge from those AUROC implies~\cite{vandervegt_threshold_sepsis_2026}. Each remains a quantity averaged over patients. Closest to our concern is CAREBench~\cite{keoliya2025carebench}, which elevates the temporal stability of an individual risk trajectory to a primary metric. That stability is within-patient. Ours is between-pipeline, over the set of patients each one flags.

\section{Problem Formulation}
\label{sec:problem}

What reaches a clinician is a list of names, not a column of scores. That difference decides what a comparison between imputation strategies has to measure.

Consider $N$ ICU encounters. Encounter $i$ contributes an hourly series $x_i\in\mathbb{R}^{T_i\times d}$ of $d$ clinical variables, an observation mask $m_i\in\{0,1\}^{T_i\times d}$, and a binary ever-septic label $y_i$. An imputation strategy $a$ is a map $\mathcal{I}_a:(x_i,m_i)\mapsto\tilde{x}_i^{(a)}$ that fills missing entries and leaves observed ones untouched. Strategies therefore differ only where data are absent. For each strategy $a$, a classifier $f_{\theta_a}$ is fitted separately
using the same model configuration. Its predictions are calibrated
and aggregated to obtain patient-level risk $p_i^{(a)}$.
At alert rate $k$ the deployment acts on the induced set
\[
A_k^{(a)}=\bigl\{\,i:\ p_i^{(a)} \text{ among the top } 100k\% \,\bigr\},
\]
which standardizes alert burden across pipelines~\cite{vandervegt_threshold_sepsis_2026}. Two strategies are operationally equivalent when they induce the same set. The natural measure of that agreement is the Jaccard index, a standard
set-overlap statistic, so what is new here is the protocol of applying it
to top-$k$ alert sets rather than the statistic itself,
\[
J_k(a,b)=\frac{|A_k^{(a)}\cap A_k^{(b)}|}{|A_k^{(a)}\cup A_k^{(b)}|}\in[0,1],
\]
equal to 1 when two pipelines flag identical patients and 0 when they share none.

AUROC cannot stand in for $J_k$, and the reason lies in what each one is sensitive to. AUROC is the probability that a randomly drawn positive outscores a randomly drawn negative,
\[
\mathrm{AUROC}=\tfrac{1}{|P||Q|}\textstyle\sum_{i\in P}\sum_{j\in Q}\mathbf{1}\!\left[p_i>p_j\right],
\]
with $P$ and $Q$ the septic and non-septic encounters. It depends only on how the two classes interleave. Any permutation of scores \emph{within} $Q$ therefore leaves it exactly unchanged. The alert list has no such invariance. Which patients occupy the top $100k\%$ turns on the ordering among non-septic encounters, and at a patient-level prevalence of 7.3\% those encounters are roughly seven in ten of the list at $k=10\%$. A rearrangement invisible to AUROC can thus replace most of $A_k^{(a)}$.

Two pipelines can therefore score identically on AUROC while sharing few of the same flagged patients. Aggregate discrimination places no floor on $J_k(a,b)$. How large that slack is in practice is an empirical question, and it is the one we take up. Our primary quantity is $J_k(a,b)$ across all method pairs at $k\in\{5,10,15,20\}\%$. Discrimination, calibration, operating-point utility, computational cost, and the clinical characteristics of the patients on whom strategies disagree are secondary.

\section{Methods}
\label{sec:methods}

Every condition runs through the same four-stage pipeline: impute, fit one fixed classifier, calibrate, then rank patients and cut at the alert rate. Only the first stage varies. Everything downstream is held constant, and everything upstream is estimated without touching the test set, so any difference in the resulting alert lists is attributable to imputation alone.

\subsection{Data, Ethics, and Split} \label{sec:data}
We use the public PhysioNet/CinC 2019 challenge data (version 1.0.0, CC~BY~4.0)~\cite{physionet_challenge_2019}. These are hourly ICU records from two U.S.\ hospital systems, Beth Israel Deaconess Medical Center (set A) and Emory University Hospital (set B), labeled with Sepsis-3 criteria. Each row carries 40 variables, plus the label: 8 vital signs, 26 laboratory values and 6 demographic or admission fields. The organizers collected the records with approval of the respective institutional review boards and de-identified them before release~\cite{reyna_physionet_sepsis_2020}. This secondary analysis of public de-identified data required no additional ethics approval or consent.

The cohort comprises 40,336 patients and 1,552,210 patient-hours. Missingness averages 71.9\% across the 39 model features. Of the 26 laboratory variables, 25 are absent from more than 90\% of patient-hours; glucose, the exception, from 83\%. The gaps are long as well as frequent: across 1,475,260 within-patient intervals between consecutive laboratory measurements, the median is 5~h (IQR 2 to 11~h), and the longest is 326~h. That profile is what clinically selective ordering looks like, and it is the regime the comparison is meant to probe.

We analyze every patient in the public release; none were excluded. Patients were split 70/30 into training (28,235) and held-out test (12,101) sets, stratified on the ever-septic label, so no patient contributes rows to both. All imputation, classifier and calibration parameters were estimated on the training set alone.

\subsection{Imputation Strategies}
We evaluate nine imputers and one native-handling baseline (\tabref{tab:agg}), spanning heuristics to deep architectures. Three constraints are shared by all: parameters fitted on training data only, causal temporal ordering, and a row-level formulation matching the classifier.

XGBoostNative is a reference condition rather than an imputer. It passes the incomplete matrix straight to the learner, whose sparsity-aware split finding learns a default direction for missing values at each node~\cite{chen_xgboost_2016}. ForwardFill (LOCF) propagates the last observation within an encounter after sorting on (patient, hour), falling back to training means before the first observation. Mean imputation uses training-set marginal means. MICE runs 10 chained-equation iterations with a Bayesian ridge estimator~\cite{buuren_mice_2018}; we pass one imputed dataset downstream so that every strategy shares an architecture (\secref{sec:limits}). The autoencoders differ only in their corruption: DAE reconstructs clean vectors from Gaussian-corrupted inputs~\cite{vincent_denoising_autoencoders_2010}, MAE masks and reconstructs observed values~\cite{he_mae_2022}, and MIDA averages $K=5$ DAE draws~\cite{gondara_mida_2018}. Three sequential imputers follow. SAITS uses self-attention~\cite{du_saits_2023}, BRITS is bidirectional recurrent~\cite{cao2018brits}, and GRU-D decays toward training means with time since the last observation~\cite{che_grud_2018}.

The neural imputers share a training budget. It is 20 epochs, hidden width 128, latent width 64, batch size 4,096, Adam at $10^{-3}$, and dropout 0.1, raised to 0.3 for GRU-D, on features standardized with training statistics. Gaussian noise has standard deviation 0.1 for DAE and MIDA, the MAE mask ratio is 0.2, and SAITS uses 4 heads over 2 layers. Reconstruction loss is computed on observed entries only, and observed values are restored afterwards. The imputed matrices are therefore identical wherever data were observed.

One design choice needs to be stated explicitly. We implement SAITS, BRITS and GRU-D per patient-hour rather than over whole trajectories. A full-trajectory imputer feeding a row-level classifier would capture temporal structure that the classifier then discards, which would confound imputation and classifier effects. The cost is real: the central strength of a sequential imputer goes unexercised, and our results characterize these methods paired with a row-level classifier, nothing more.

\subsection{Classifier and Calibration}
Every imputed matrix feeds the same XGBoost classifier~\cite{chen_xgboost_2016}, predicting the hourly Sepsis-3 label from 39 features. The 40th variable is the hour index, which orders rows but never enters as a feature. Settings are 100 trees, maximum depth 5, learning rate 0.03, row and column subsampling 0.8, minimum child weight 5, gamma 1, histogram tree method, and positive-class weight 54.5, computed as the training negative-to-positive row ratio. They were fixed before the comparison and applied unchanged to every strategy, with no per-method tuning. Training ran the full 100 rounds; no early stopping was used and no model selection depended on test-set performance.

Raw scores are Platt-calibrated~\cite{platt_scaling_1999}, a sigmoid fitted on a stratified 20\% subsample of the same training rows. Platt scaling is monotone, so it changes no ranking: alert lists, Jaccard, AUROC, precision and recall are identical before and after it. Brier and ECE do depend on it, and since the calibration sample is in-sample they are optimistic in absolute terms; we use them only to compare methods, which share the procedure. Per-patient risk is the maximum calibrated probability over an encounter's hours, mirroring an alert that fires the first time hourly risk crosses the threshold. Alternative aggregation rules are examined in \secref{sec:agg}.

\subsection{Outcomes and Analysis}
The primary outcome is alert-list concordance. For each method the top $k$\% are the test patients scoring at or above the $(1-k)$ quantile of that method's patient-level scores, for $k\in\{5,10,15,20\}$, ties retained. We compare all 45 pairs at every threshold. Confidence intervals come from 1,000 bootstrap resamples of the paired flag vectors, drawn at the patient level, the unit at which alerts are issued. We also report Spearman and Kendall rank correlation between patient-level score vectors, to separate agreement over the whole ranking from agreement at the cut. Alongside these we compute AUROC, AUPRC, Brier score, and ECE over ten uniform bins on the held-out test set, together with precision and recall at each $k$ and total pipeline time.

Three checks probe whether the concordance structure is an artifact of how we cut or aggregate. We recompute Jaccard at $k\in\{6,\dots,12\}$, and again within each hospital system with thresholds set per site. We then vary the per-patient aggregation rule five ways: mean of the top three hours, 90th percentile, mean of the last six hours, maximum over the first 24~h, and mean over the stay. Patient-level AUROC and AUPRC use 1,000 outcome-stratified bootstrap resamples, with pairwise differences tested by DeLong's method. All tests are two-sided at $\alpha=0.05$, Bonferroni-corrected within each family of comparisons.

To characterize where the methods disagree, we take the 880 sepsis-positive test encounters and assign them at $k=10\%$, for three pairs spanning the observed range of concordance, to four groups: concordant high-risk, only-A, only-B, and concordant low-risk. Two contrast ForwardFill with a reconstruction imputer ($J=0.424$ and $0.446$); one contrasts two reconstruction imputers ($0.606$). The full matrix runs 0.399 to 0.750. ICU length of stay is heavily right-skewed, so we compare it across groups with the Kruskal-Wallis test and Dunn's post-hoc test under Bonferroni correction.

Experiments ran on one node with an NVIDIA V100 GPU (24~GB), 4 CPU cores and 64~GB RAM. Neural imputers used the GPU and the rest ran on CPU, under Python 3.9, XGBoost 1.7.0, scikit-learn 1.0.2 and PyTorch 2.0.0, with seed 42 throughout.

\section{Results}
\label{sec:results}

Every number below comes from the same 12,101 held-out patients, scored by the same classifier and calibrated the same way. A difference between two rows is therefore a difference between imputers and nothing else. We report the aggregate metrics first, because those are what current practice would use to choose among these methods. The alert lists come next, then what those lists deliver at the operating point, then three robustness checks. We close with the clinical profile of the disagreement and its computational cost.

\subsection{Aggregate Discrimination and Calibration}
A conventional comparison would stop at the metrics in \tabref{tab:agg}, which we treat as secondary. Eight of ten strategies cluster within an AUROC band of 0.011, with ForwardFill and XGBoostNative clearly above them. That band is narrower than the patient-level bootstrap interval of any single method. Brier scores are effectively identical, and AUPRC runs from 2.7 to 3.8 times the 0.0179 expected at random.

Calibration behaves as expected under this much imbalance. Every method had ECE above 0.37 before calibration; after Platt scaling, row-level ECE fell below 0.002 for all ten. That figure should not be over-read. Calibrated probabilities never exceed 0.27, so only two or three of the ten uniform bins are ever occupied. Patient-level calibration was similar across methods as well. Brier ran 0.063 to 0.064 and ten-bin ECE 0.027 to 0.040, with every pipeline under-predicting ever-septic risk to a comparable degree: mean predicted risk 0.033 to 0.045, against a prevalence of 0.073. Read carefully, these numbers say only that no method is grossly miscalibrated relative to the others. A ten-bin ECE is insensitive by construction when predictions sit in so narrow a range, and a genuine reliability assessment would need bins on predicted quantiles instead. If evaluation stopped here, almost any reconstruction-based imputer would look acceptable.

\begin{table}[t]
\centering
\caption{Row-level aggregate metrics on the held-out test set, sorted by AUROC, with total pipeline time. ECE is post-Platt.}
\label{tab:agg}
\footnotesize
\setlength{\tabcolsep}{2.2pt}
\resizebox{\linewidth}{!}{\begin{tabular}{@{}lccccc@{}}
\toprule
Method & AUROC & AUPRC & Brier & ECE & Time (min) \\
\midrule
ForwardFill & 0.787 & 0.068 & 0.01718 & 0.00172 & 0.69 \\
XGBoostNative$^\ddagger$ & 0.744 & 0.059 & 0.01728 & 0.00071 & 90.13 \\
Mean & 0.733 & 0.054 & 0.01733 & 0.00064 & 0.62 \\
SAITS & 0.732 & 0.052 & 0.01734 & 0.00048 & 7.54 \\
BRITS & 0.732 & 0.052 & 0.01733 & 0.00032 & 3.18 \\
DAE & 0.730 & 0.051 & 0.01735 & 0.00047 & 1.63 \\
MICE$^\S$ & 0.728 & 0.051 & 0.01734 & 0.00005 & 5.78 \\
MIDA & 0.727 & 0.050 & 0.01736 & 0.00040 & 1.85 \\
GRU-D & 0.724 & 0.049 & 0.01737 & 0.00029 & 3.52 \\
MAE & 0.722 & 0.049 & 0.01737 & 0.00030 & 1.51 \\
\bottomrule
\end{tabular}}
\\[2pt]{\footnotesize $^\ddagger$Native-handling baseline, not an imputation strategy; its time is tree growth, not imputation. $^\S$One imputed draw, so this row describes single-draw MICE rather than MICE under Rubin's rules.}
\end{table}

\subsection{Patient-Level Ranking Stability}
\begin{figure}[t]
\centering
\includegraphics[width=\linewidth]{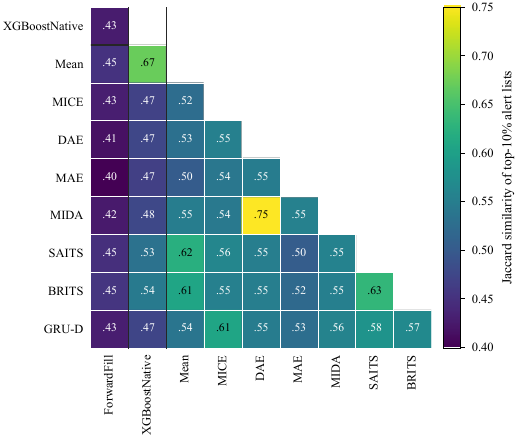}
\caption{Pairwise Jaccard of top-10\% alert lists, lower triangle (the matrix is symmetric). Rules separate the three tiers: ForwardFill, the native-handling baseline, and the eight reconstruction imputers.}
\label{fig:heatmap}
\end{figure}

We now turn to the primary outcome. Given the same alert budget, do these methods select the same patients? \figref{fig:heatmap} answers that for all 45 pairs at $k=10\%$. Every method flagged exactly 1,211 patients, so no cutoff ties arose. ForwardFill is a persistent outlier. Its values fall between 0.399 and 0.453, below every other pair in the matrix. Put concretely: in a 100-bed ICU at a 10\% alert rate, ForwardFill and any reconstruction-based method would differ on about 4 of the 10 patients on the top-decile list. Reconstruction-based pairs range from 0.502 to 0.750, the highest being DAE and MIDA (0.750, 95\% CI 0.726 to 0.771). The native-handling baseline sits between the two groups, agreeing with Mean at 0.670 (0.645 to 0.696) but with the other imputers at only 0.465 to 0.543. Bootstrap intervals have mean width 0.048 and maximum 0.051. High agreement across the full ranking does not guarantee agreement among the patients near the threshold. Spearman correlation between patient-level scores runs 0.825 to 0.975 across the 45 pairs, Kendall's $\tau$ 0.633 to 0.872. Top-decile Jaccard for those same pairs runs 0.399 to 0.750. DAE and MIDA correlate at 0.975 and still differ on a quarter of the union of their lists. Agreement over the whole ranking is compatible with disagreement at the threshold, since only the ordering near the cut decides membership.

Because alert-list overlap may change with the alert rate, we repeated the analysis at several thresholds. Jaccard broadly increases with $k$ for every pair (\tabref{tab:thresh}). The separation between the ForwardFill and reconstruction-only blocks holds at every operating point, including each step of the finer grid from $k=6$ to $12\%$. The gap is widest at $k=5\%$, where ForwardFill and MAE have a Jaccard overlap of 0.305. The methods therefore disagree most when the alert budget is smallest.

\begin{table}[t]
\centering
\caption{Jaccard for ten representative pairs across alert rates (pooled), and at $k=10\%$ within each hospital system with thresholds set per site (A: $n=6{,}118$; B: $n=5{,}983$).}
\label{tab:thresh}
\scriptsize
\setlength{\tabcolsep}{1.9pt}
\begin{tabular}{@{}lcccccccc@{}}
\toprule
& \multicolumn{6}{c}{Pooled, $k$ (\%)} & \multicolumn{2}{c}{$k{=}10\%$ by site}\\
\cmidrule(lr){2-7}\cmidrule(lr){8-9}
Method pair & 5 & 8 & 10 & 12 & 15 & 20 & A & B \\
\midrule
DAE-MIDA & 0.679 & 0.741 & 0.750 & 0.785 & 0.787 & 0.808 & 0.707 & 0.744 \\
Mean-XGBNative & 0.584 & 0.651 & 0.670 & 0.703 & 0.703 & 0.750 & 0.645 & 0.683 \\
BRITS-Mean & 0.554 & 0.607 & 0.606 & 0.646 & 0.656 & 0.700 & 0.602 & 0.593 \\
BRITS-SAITS & 0.540 & 0.607 & 0.631 & 0.633 & 0.649 & 0.679 & 0.569 & 0.610 \\
GRU-D-MICE & 0.538 & 0.571 & 0.605 & 0.628 & 0.658 & 0.678 & 0.545 & 0.589 \\
Mean-SAITS & 0.530 & 0.589 & 0.620 & 0.643 & 0.662 & 0.699 & 0.592 & 0.632 \\
\midrule
FF-SAITS & 0.341 & 0.419 & 0.446 & 0.467 & 0.516 & 0.547 & 0.409 & 0.431 \\
FF-Mean & 0.333 & 0.404 & 0.451 & 0.488 & 0.534 & 0.569 & 0.402 & 0.440 \\
FF-MIDA & 0.315 & 0.375 & 0.424 & 0.441 & 0.485 & 0.524 & 0.362 & 0.424 \\
FF-MAE & 0.305 & 0.374 & 0.399 & 0.419 & 0.469 & 0.519 & 0.368 & 0.393 \\
\bottomrule
\end{tabular}
\end{table}

\begin{figure*}[!t]
\centering
\includegraphics[width=\linewidth]{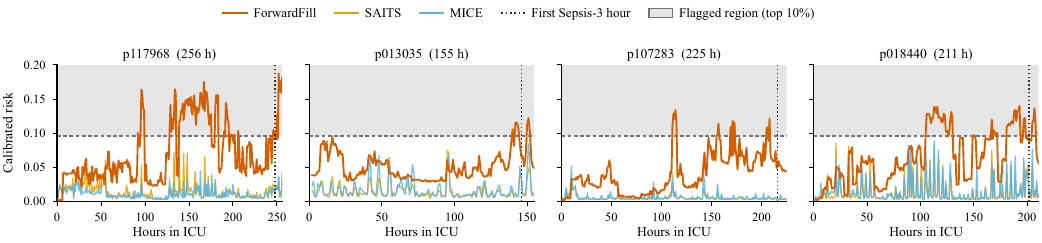}
\caption{Hourly calibrated risk in four only-ForwardFill encounters: septic patients ForwardFill flags at $k=10\%$ while SAITS and MICE do not. Dotted line: first Sepsis-3 labeled hour. Shaded band: the flagged region. Panels share a vertical scale.}
\label{fig:traj}
\end{figure*}

\subsection{Operating-Point Utility}
Patient-level precision and recall at all four alert rates appear in \tabref{tab:ppv}. At $k=10\%$ precision spanned 0.257 to 0.298, roughly four times the 0.073 of a random patient list, with recall from 0.353 to 0.410. The other thresholds behave the same way: precision falls from 0.294--0.371 at $k=5\%$ to 0.207--0.225 at $k=20\%$ as recall rises. ForwardFill has the highest precision among imputers at every threshold, in line with its AUROC. Yet the spread across all ten methods at $k=10\%$ is only 0.041, an order of magnitude smaller than the disagreement in list membership. Utility at a fixed rate is not blind to these differences, but it sees a compressed version of the same ordinal signal AUROC reports, with the three leading methods leading here too. Jaccard instead reflects which patients make up the list.

\begin{table}[t]
\centering
\caption{Patient-level precision@$k$ and recall@$k$ on the held-out test set ($n=12{,}101$ patients, 880 ever-septic).}
\label{tab:ppv}
\scriptsize
\setlength{\tabcolsep}{2.6pt}
\begin{tabular}{@{}lcccccccc@{}}
\toprule
& \multicolumn{2}{c}{$k{=}5\%$} & \multicolumn{2}{c}{$k{=}10\%$} & \multicolumn{2}{c}{$k{=}15\%$} & \multicolumn{2}{c}{$k{=}20\%$}\\
\cmidrule(lr){2-3}\cmidrule(lr){4-5}\cmidrule(lr){6-7}\cmidrule(lr){8-9}
Method & Prec & Rec & Prec & Rec & Prec & Rec & Prec & Rec \\
\midrule
ForwardFill & 0.371 & 0.256 & 0.297 & 0.409 & 0.253 & 0.522 & 0.225 & 0.619 \\
XGBoostNative$^\ddagger$ & 0.347 & 0.239 & 0.298 & 0.410 & 0.249 & 0.515 & 0.224 & 0.616 \\
Mean & 0.340 & 0.234 & 0.291 & 0.400 & 0.247 & 0.510 & 0.220 & 0.605 \\
SAITS & 0.328 & 0.226 & 0.285 & 0.392 & 0.248 & 0.512 & 0.220 & 0.605 \\
BRITS & 0.314 & 0.216 & 0.273 & 0.375 & 0.235 & 0.484 & 0.210 & 0.578 \\
MICE & 0.317 & 0.218 & 0.269 & 0.370 & 0.235 & 0.485 & 0.211 & 0.582 \\
MIDA & 0.314 & 0.216 & 0.267 & 0.367 & 0.231 & 0.476 & 0.211 & 0.581 \\
DAE & 0.300 & 0.207 & 0.269 & 0.370 & 0.234 & 0.483 & 0.211 & 0.580 \\
GRU-D & 0.309 & 0.212 & 0.257 & 0.353 & 0.229 & 0.472 & 0.207 & 0.569 \\
MAE & 0.294 & 0.202 & 0.259 & 0.357 & 0.229 & 0.472 & 0.207 & 0.570 \\
\bottomrule
\end{tabular}
\\[2pt]{\footnotesize $^\ddagger$Native-handling baseline, not an imputation strategy.}
\end{table}

\subsection{Site-Wise and Patient-Level Robustness}
The pooled test set mixes two hospital systems, so we repeat the analysis within each (\tabref{tab:thresh}, last two columns; \tabref{tab:ptauroc}). The same pattern appears in both hospital systems. ForwardFill's Jaccard with every other method reaches at most 0.409 in site A and 0.440 in site B, while all other pairs begin at 0.466 and 0.454. The two groups do not overlap in either site. Patient-level AUROC runs higher than the row-level values, as expected with one score per encounter. The broad performance pattern persists across sites, and the eight reconstruction-based methods span 0.786 to 0.804 with overlapping intervals. DeLong tests found no difference between ForwardFill and XGBoostNative ($p=0.35$). Against the reconstruction methods, ForwardFill is higher in all eight comparisons, and seven survive Bonferroni correction with differences of 0.014 to 0.030. Mean is the exception at 0.012 ($p=0.008$ against a corrected threshold of 0.006). It is the only reconstruction method whose discrimination is not significantly below ForwardFill's. Among the 28 reconstruction-method pairs, only 6 reach significance, with AUROC differences no larger than 0.018. The remaining 22 pairs illustrate the central result without a meaningful difference in discrimination. Their AUROC differences are at most 0.011 and none survives Bonferroni correction, yet their alert lists overlap at Jaccard 0.518 to 0.750. MAE and BRITS are the sharpest case: 0.010 apart in AUROC, indistinguishable under test, and a Jaccard of 0.518. Comparable aggregate discrimination therefore did not imply that two pipelines nominated the same patients.

\begin{table}[t]
\centering
\caption{Patient-level discrimination on the maximum calibrated risk: pooled AUROC with 1,000-resample stratified bootstrap 95\% CI, AUPRC, and AUROC within each hospital system.}
\label{tab:ptauroc}
\scriptsize
\setlength{\tabcolsep}{2.2pt}
\begin{tabular}{@{}lccccc@{}}
\toprule
Method & AUROC & 95\% CI & AUPRC & Site A & Site B \\
\midrule
ForwardFill & 0.816 & 0.803--0.830 & 0.292 & 0.801 & 0.830 \\
XGBoostNative$^\ddagger$ & 0.812 & 0.798--0.826 & 0.274 & 0.801 & 0.818 \\
Mean & 0.804 & 0.790--0.819 & 0.252 & 0.788 & 0.814 \\
SAITS & 0.802 & 0.788--0.817 & 0.252 & 0.791 & 0.809 \\
BRITS & 0.797 & 0.784--0.812 & 0.241 & 0.780 & 0.810 \\
MICE & 0.797 & 0.782--0.812 & 0.239 & 0.785 & 0.809 \\
DAE & 0.794 & 0.780--0.809 & 0.233 & 0.782 & 0.806 \\
MIDA & 0.794 & 0.779--0.809 & 0.241 & 0.784 & 0.801 \\
MAE & 0.788 & 0.774--0.803 & 0.233 & 0.776 & 0.797 \\
GRU-D & 0.786 & 0.772--0.801 & 0.233 & 0.774 & 0.799 \\
\bottomrule
\end{tabular}
\\[2pt]{\scriptsize $^\ddagger$Native-handling baseline. Site columns are AUROC; A $n=6{,}118$ (531 septic), B $n=5{,}983$ (349 septic).}
\end{table}

\subsection{Sensitivity to the Per-Patient Aggregation Rule}
\label{sec:agg}
Maximum-over-stay aggregation mechanically favors long stays, so we recompute patient-level scores under five other rules (\tabref{tab:agg_rule}). The ForwardFill outlier is insensitive to the rule, staying between 0.36 and 0.46 under all six while reconstruction-only pairs range from 0.45 to 0.90. The length-of-stay pattern, by contrast, does depend on the rule. Under whole-stay rules the only-ForwardFill group is long-stay and the only-MIDA group short-stay. Switch to a first-24-hour window and the four groups have similar median stays, 33 to 39~h.

\begin{table}[t]
\centering
\caption{Sensitivity to the per-patient aggregation rule at $k=10\%$. Jaccard ranges cover the nine FF pairs and the 36 pairs among the nine non-ForwardFill methods, including XGBoostNative; LOS is the median for septic patients flagged by only one method.}
\label{tab:agg_rule}
\scriptsize
\setlength{\tabcolsep}{2.2pt}
\resizebox{\linewidth}{!}{\begin{tabular}{@{}lcccc@{}}
\toprule
Rule & FF vs others & Other pairs (med.) & LOS FF/MIDA (h) & AUROC recon. \\
\midrule
Max over stay & 0.40--0.45 & 0.47--0.75 (0.55) & 55 / 43 & 0.786--0.804 \\
Mean of top 3~h & 0.42--0.46 & 0.53--0.86 (0.63) & 47 / 46 & 0.788--0.800 \\
90th percentile & 0.39--0.43 & 0.53--0.84 (0.65) & 65 / 23 & 0.790--0.803 \\
Mean of last 6~h & 0.36--0.40 & 0.59--0.88 (0.71) & 61 / 22 & 0.816--0.828 \\
Max, first 24~h & 0.40--0.46 & 0.45--0.75 (0.53) & 33 / 39 & 0.740--0.763 \\
Mean over stay & 0.38--0.41 & 0.59--0.90 (0.73) & 72 / 20 & 0.782--0.795 \\
\bottomrule
\end{tabular}}
\end{table}

\subsection{Clinical Correlates of Disagreement}
To judge whether the disagreement matters, we examined the patients it falls on. \tabref{tab:subgroup} reports ICU LOS and method-specific risk for the four agreement groups in three pairs. Concordant high-risk encounters had the longest stays, concordant low-risk the shortest. Encounters flagged only by ForwardFill had stays close to the concordant high-risk group, yet their reconstructed risks sat far below it. The converse group showed the opposite: shorter stays, and low ForwardFill risk. Kruskal-Wallis tests were significant in every panel ($H(3)=155.97$, $159.87$, $94.22$; all $p<0.001$). Dunn tests place the concordant low-risk group below every other group in Panels A and B; in Panel C only its contrast with the both-flagged group survived correction.

The same table yields a positive-case Jaccard, the overlap of flagged septic encounters: 0.574, 0.570 and 0.722. Each exceeds the all-patient Jaccard of the same pair (0.424, 0.446, 0.606), so disagreement concentrates among non-septic patients. Uniquely flagged patients are also alerted late. ForwardFill's first threshold crossing preceded the first labeled hour in 63 of 111 only-ForwardFill patients, a median lead of 5~h, and MIDA's in 41 of 74 only-MIDA patients, 2.5~h. Concordant patients had leads of 22 to 26~h. \figref{fig:traj} traces the mechanism hour by hour. Of the 12 only-ForwardFill encounters staying at least 150~h, it shows the four where ForwardFill exceeds both reconstruction methods for the largest fraction of hours. These are an illustrative best case for the mechanism, not a representative sample of the 111 only-ForwardFill encounters.

\begin{table}[t]
\centering
\caption{Agreement groups among the 880 sepsis-positive test encounters at $k=10\%$. ICU LOS in hours as median [IQR] (mean $\pm$ SD); risks are calibrated probabilities.}
\label{tab:subgroup}
\scriptsize
\setlength{\tabcolsep}{2.5pt}
\begin{tabular}{@{}lrlcc@{}}
\toprule
Group & $n$ & ICU LOS (h) & Risk A & Risk B \\
\midrule
\multicolumn{5}{@{}l}{\textit{Panel A: ForwardFill (A) vs.\ MIDA (B)}}\\
Both flagged & 249 & 75.0 [30.0, 130.0] (91.1 $\pm$ 72.4) & 0.152 & 0.129 \\
Only ForwardFill & 111 & 55.0 [27.0, 102.0] (76.1 $\pm$ 64.8) & 0.132 & 0.078 \\
Only MIDA & 74 & 43.0 [18.0, 79.5] (58.3 $\pm$ 54.7) & 0.067 & 0.116 \\
Neither & 446 & 22.0 [11.0, 46.0] (37.7 $\pm$ 42.4) & 0.040 & 0.054 \\
\midrule
\multicolumn{5}{@{}l}{\textit{Panel B: ForwardFill (A) vs.\ SAITS (B)}}\\
Both flagged & 256 & 71.5 [30.0, 123.0] (87.3 $\pm$ 69.2) & 0.153 & 0.133 \\
Only ForwardFill & 104 & 61.0 [26.0, 119.5] (84.5 $\pm$ 73.6) & 0.129 & 0.078 \\
Only SAITS & 89 & 43.0 [18.0, 86.0] (56.7 $\pm$ 45.2) & 0.066 & 0.118 \\
Neither & 431 & 21.0 [11.0, 45.5] (37.3 $\pm$ 44.1) & 0.039 & 0.053 \\
\midrule
\multicolumn{5}{@{}l}{\textit{Panel C: Mean (A) vs.\ BRITS (B)}}\\
Both flagged & 286 & 61.5 [27.0, 114.0] (81.0 $\pm$ 67.7) & 0.144 & 0.129 \\
Only Mean & 66 & 48.5 [21.2, 93.8] (70.4 $\pm$ 64.9) & 0.116 & 0.085 \\
Only BRITS & 44 & 42.0 [20.8, 80.8] (68.5 $\pm$ 71.1) & 0.085 & 0.109 \\
Neither & 484 & 23.0 [11.8, 55.0] (44.3 $\pm$ 50.3) & 0.053 & 0.054 \\
\bottomrule
\end{tabular}
\end{table}

The same structure appears in the AUROC-indistinguishable pair. For MAE and BRITS the four groups differ in stay (Kruskal-Wallis $H(3)=93.6$, $p<0.001$), concordant high-risk longest at 68~h and concordant low-risk shortest at 23~h. Positive-case Jaccard is 0.610 against an all-patient 0.518. Uniquely flagged patients are again alerted late, a median 6.5~h for only-MAE and 1.0~h for only-BRITS, against 21~h for concordant patients. The clinical signature of disagreement is therefore not specific to ForwardFill.

In each panel of \figref{fig:traj}, ForwardFill runs above both reconstruction methods for more than 90\% of hours, and its peak crosses the alert threshold. SAITS and MICE fall back toward baseline over the same intervals and never cross theirs. This is the pattern expected if ForwardFill carries an earlier abnormal value through the hours when no new measurement is recorded.

\subsection{Computational Cost}
Total pipeline time per method is listed in \tabref{tab:agg}. The autoencoders are the efficient neural choice. DAE, MAE and MIDA finish in 1.5 to 1.9~min, against 3.2~min for BRITS and 7.5~min for SAITS, with row-level AUROC within 0.010 of both and Jaccard in the same tier. These are wall-clock times on heterogeneous hardware, neural imputers on GPU and the rest on CPU, so read them as orders of magnitude rather than exact ratios. Any bias from this favors the deep imputers, and a CPU-only rerun would settle the comparison. The sequential imputers cost 5 to 12 times Mean while enjoying GPU acceleration Mean did not, so the gap would widen on uniform hardware; the autoencoders cost 2.4 to 3.0 times. XGBoostNative's 90.13~min is tree growth, not imputation.

\section{Discussion}
\label{sec:discussion}

Eight strategies sit within 0.011 AUROC of one another, yet their alert lists overlap at Jaccard 0.50 to 0.75. Twenty-two of those pairs cannot be statistically distinguished on AUROC and still disagree. This is the direct evidence for the paper's central claim. Population-level discrimination averages error over all patients. It says nothing about whether two pipelines flag the same people at a fixed operating point. Outcome studies, equity analyses~\cite{getzen2023equitable} and audits spanning a pipeline change therefore inherit the imputation choice, even when AUROC and Brier are unchanged. ForwardFill shows a related but distinct pattern. It pairs higher discrimination with the lowest overlap in the matrix, so its divergence cannot be read as equal discrimination producing unequal alerts.

The disagreement follows a systematic pattern. DAE and MIDA share an encoder-decoder and differ only in noise draws, and they agree most. Mean and XGBoostNative both leave frequently observed values untouched, and they come next. Methods from different families agree near 0.50 despite AUROC differences inside that same 0.011 band. How a method fills the missing entries appears to govern which patients reach the list. The small differences in discrimination between these methods explain much less.

ForwardFill's behavior fits the likely MNAR structure of ICU laboratory data. Tests are ordered on suspicion of deterioration, so the last observed value is systematically more abnormal than a population-centered reconstruction. LOCF carries that value forward and amplifies a signal the reconstruction methods attenuate. This agrees with Sharafoddini et al., who found the missingness indicator alone prognostic~\cite{sharafoddini_missing_icu_2019}. It contrasts with evaluation practice that prioritizes reconstruction accuracy: Poette et al.\ rank SAITS, with US-GAN, as the most accurate reconstructors on MIMIC-IV~\cite{poette2026benchmarking}, yet here SAITS neither led discrimination nor produced a distinctive alert list. Reconstruction fidelity and operational behavior are different objectives. We did not directly test this explanation, so it remains an interpretation.

There is also an important alternative explanation. LOCF may simply propagate stale extreme values, so a patient with one abnormal early measurement and a long silence accumulates risk whether or not they are deteriorating. The lead times support that caution. Patients flagged only by ForwardFill are alerted a median of 5~h before the first labeled hour, against 22 to 26~h for concordant patients. Its unique detections are late, not early. Concordance analysis alone cannot settle whether they are genuine sensitivity or artifact.

That the deep imputers did not beat Mean holds for row-level pipelines only, and three factors compound. The row-level formulation discards the temporal dependencies SAITS, BRITS and GRU-D are built to exploit. At 71.9\% missingness, reconstruction must infer most of each vector from a minority of observed entries. And under MNAR the absence pattern is itself informative, so better reconstruction need not improve discrimination~\cite{lipton2016missing}.

Disagreement concentrates among non-septic patients. Positive-case Jaccard exceeds all-patient Jaccard for every pair, and precision varies by at most 0.041. Switching pipelines therefore mostly exchanges one set of false positives for another rather than changing how many true cases are caught. Sensitivity changes little, but clinicians may still face a different set of false-positive alerts. The deployment consequence is that a pipeline change moves patients on and off the list with no visible change in AUROC, so such changes should be evaluated by overlap with the prior deployment. For row-level tree-based pipelines under extreme missingness, Mean imputation is an efficient baseline: fastest, and the only reconstruction method whose AUROC is not significantly below ForwardFill's. Its high agreement with the other reconstruction methods places it inside that cluster. It does not show that its alert list is the clinically better one. Jaccard measures how stable a decision set is under a preprocessing change, not which decision set is right.

\section{Limitations}
\label{sec:limits}

Our results come from one cohort of two U.S.\ hospital systems. The site-wise analysis shows the findings hold within each, but the models are trained on pooled data, so a transfer test remains open. External validation on a cohort such as MIMIC-IV would show whether the tier structure belongs to this cohort or to the task.

The row-level implementation of SAITS, BRITS and GRU-D is the most important limit on what these results cover. Pairing them in full-trajectory form with a sequential classifier would test whether sequential inductive bias yields a patient-level advantage. We evaluate MICE on one imputed dataset and MIDA on the mean of five draws, so imputation uncertainty is not propagated downstream and the MICE result describes that formulation rather than deployed MICE under Rubin's rules. We use one learner family throughout, and per-method tuning was excluded by design, so tuning each imputer separately could in principle reorder the tiers.

We exclude missingness indicators and time-since-observation features to isolate the imputed values. Adding them is well supported~\cite{lipton2016missing,che_grud_2018,vanness2023mim,sisk2023indicators}, and their absence may disadvantage the imputed pipelines relative to XGBoostNative, which sees the pattern natively. We study a single tabular modality, and missingness in multimodal clinical pipelines raises further questions that we do not address here~\cite{raju2025multimodal}. Finally, we show that imputation changes who is flagged, not that this alters escalation decisions or outcomes. Prospective shadow deployment is the definitive test. Code, configuration files and the pairwise concordance matrices are available from the authors on request.

\section{Conclusion}
\label{sec:conclusion}

Imputation choice can change which patients are flagged. Among 22 reconstruction-method pairs with no statistically significant patient-level AUROC difference after Bonferroni correction, AUROC differences were at most 0.011, yet top-decile alert-list Jaccard ranged from 0.518 to 0.750. ForwardFill showed a separate pattern, combining higher discrimination with Jaccard of 0.40 to 0.45 against other methods. Splitting by hospital system, or scoring patients a different way, did not change it. Which patients a clinician is asked to evaluate therefore depends on a preprocessing choice that standard reporting does not show. Alert-list concordance belongs in the evaluation of any pipeline change, alongside the aggregate metrics that cannot detect it. Under extreme missingness with a row-level tree classifier, mean imputation is an efficient baseline, chosen for cost rather than demonstrated clinical superiority. ForwardFill flags a systematically different and later-alerted cohort, and needs cohort-specific justification. The same analysis applies to any prediction system that alerts a fixed proportion of patients. It could also be computed within demographic strata as an equity check, which we did not test here.

\bibliographystyle{IEEEtran}
\bibliography{refs}

\end{document}